\documentclass[times, review, 10pt]{elsarticle}

\usepackage{amssymb}
\usepackage{amsmath}
\usepackage{booktabs}
\usepackage{tabularx}
\usepackage{threeparttable}
\usepackage{multirow}
\usepackage[normalem]{ulem}
\usepackage{makecell}
\usepackage{soul}
\usepackage{xcolor}
\sethlcolor{yellow}
\journal{Pattern Recognition}

\begin{document}

\begin{frontmatter}



\title{Palmprint De-Identification via Diffusion Model for High-Quality and Diverse Synthesis}


%\author{Licheng Yan, Bob Zhang*, Andrew Beng Jin Teoh, Lu Leng, Shuyi Li, Yuqi Wang, Ziyuan Yang} %% Author name

\author[1]{Licheng Yan}
\ead{yc48110@um.edu.mo}
\author[1]{Bob Zhang\corref{cor1}}
\ead{bobzhang@um.edu.mo}
\author[2]{Andrew Beng Jin Teoh}
\ead{bjteoh@yonsei.ac.kr}
\author[3]{Lu Leng}
\ead{leng@nchu.edu.cn}
\author[1]{Shuyi Li}
\ead{syli2022@bjut.edu.cn}
\author[1]{Yuqi Wang}
\ead{yc37500@um.edu.mo}
\author[4]{Ziyuan Yang}
\ead{cziyuanyang@gmail.com}

\cortext[cor1]{Corresponding author}

\affiliation[1]{organization={PAMI Research Group, Department of Computer and Information Science, University of Macau},%Department and Organization
            addressline={Taipa}, 
            city={Macau},
            postcode={999078}, 
%            state={Macau},
            country={China}}

\affiliation[2]{organization={School of Electrical and Electronic Engineering, College of Engineering, Yonsei University},%Department and Organization
	addressline={50 Yonsei-ro}, 
	city={Seodaemun-gu},
	postcode={120749}, 
	state={Seoul},
	country={Republic of Korea}}

\affiliation[3]{organization={School of Software, Nanchang Hangkong University},%Department and Organization
	addressline={696 Fenghe Nan Avenue}, 
	city={Nanchang},
	postcode={330063}, 
	state={Jiangxi},
	country={China}}

\affiliation[4]{organization={College of Computer Science, Sichuan University},%Department and Organization
%	addressline={696 Fenghe Nan Avenue}, 
	city={Chengdu},
	postcode={610045}, 
	state={Sichuan},
	country={China}}

%% Abstract
\begin{abstract}
Palmprint recognition techniques have advanced significantly in recent years, enabling reliable recognition even when palmprints are captured in uncontrolled or challenging environments. However, this strength also introduces new risks, as publicly available palmprint images can be misused by adversaries for malicious activities. Despite this growing concern, research on methods to obscure or anonymize palmprints remains largely unexplored. Thus, it is essential to develop a palmprint de-identification technique capable of removing identity-revealing features while retaining the image’s usability and preserving non-sensitive information. In this paper, we propose a training-free framework that utilizes pre-trained diffusion models to generate diverse, high-quality palmprint images that conceal identity features for de-identification purposes. To ensure greater stability and controllability in the synthesis process, we incorporate a \textit{semantic-guided embedding fusion} alongside a \textit{prior interpolation} mechanism. We further propose the \textit{de-identification ratio}, a novel metric for intuitive de-identification assessment. Extensive experiments across multiple palmprint datasets and recognition methods demonstrate that our method effectively conceals identity-related traits with significant diversity across de-identified samples. The de-identified samples preserve high visual fidelity and maintain excellent usability, achieving a balance between de-identification and retaining non-identity information.
\end{abstract}

%\includegraphics[width=1\textwidth]{fig_overall_pipeline}

%\item First de-identification method tailored for palmprints
%\item Training- and optimization-free de-identification framework
%\item Requires only a single hand image for de-identification
%\item Novel metric enables intuitive de-identification assessment
%\item Extensive tests on multiple datasets validate effectiveness

%% Keywords

\end{frontmatter}


\section{Introduction}\label{sec:1}
Biometrics has found widespread applications across various domains by leveraging distinctive physical or behavioral traits to represent individuals. Among these, palmprint recognition has seen rapid advancement in recent years, owing to the inherent stability and richness of palmprint features. While some studies have prioritized enhancing recognition accuracy and overall performance \cite{shen2022distribution,shao2022towards,yulin2023best,yang2023multi,yang2023comprehensive,yang2025multi}, others have addressed the challenges of recognizing palmprints captured under complex and unconstrained conditions \cite{matkowski2019palmprint,matkowski2023improving,liang2023pklnet,su2024complete}. Remarkably, even palmprints collected in uncontrolled or “in-the-wild” environments can still be effectively recognized using modern, high-capacity recognition techniques.

While recognition capabilities have advanced significantly, so too have the associated risks. Malicious actors can exploit publicly available palmprint images, often sourced from social media or publicly accessible websites, to impersonate legitimate users for malicious purposes. To mitigate such threats and safeguard the privacy of palmprints in public domains, de-identification emerges as a crucial defense. De-identification is a privacy-preserving technique that removes identity-specific information from biometric data while retaining its non-identifying features and functional usability. Although substantial progress has been made in de-identifying other biometric modalities \cite{gross2006model,li2019anonymousnet,lin2021fpgan,ghafourian2023toward,he2024diff,sun2024diffam,abd2020data,li2024hierarchical,panigrahy2009privacy}, palmprint de-identification remains largely unaddressed. Moreover, existing methods either fail to maintain the natural appearance of the de-identified data or depend on complex procedures and additional inputs. This gap underscores the urgent need for a specialized de-identification method specifically designed for palmprints.

\begin{figure}[!t]
	\centering
	\includegraphics[width=1\textwidth]{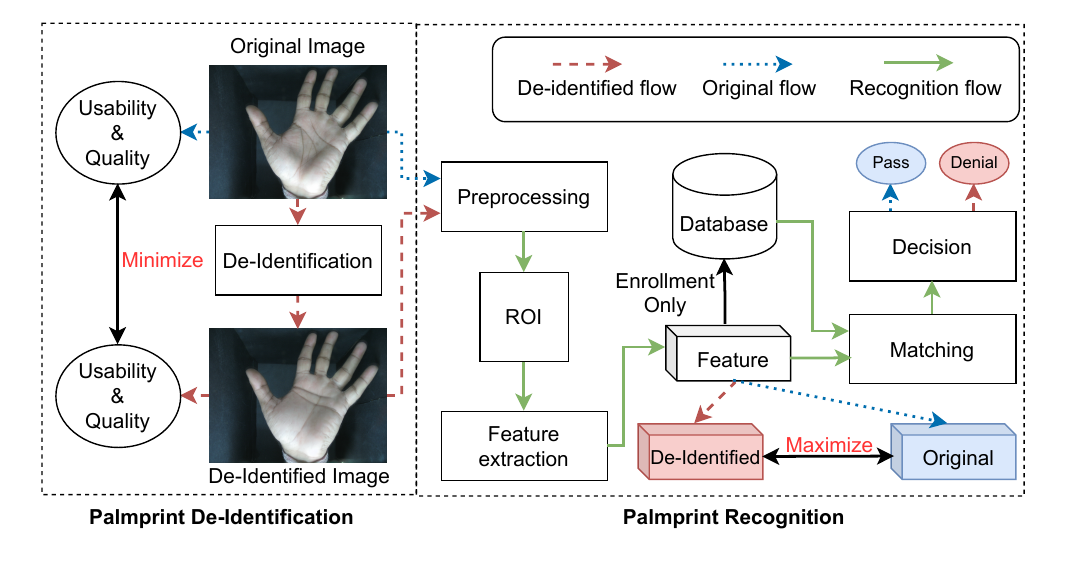}
	\caption{Palmprint de-identification and recognition pipeline. The goal of the de-identified image is to minimize the differences of usability and quality to the original and to maximize the feature space distance from the original in the recognition system for denying recognition.}
	\label{fig:pr_de-id}
\end{figure}

An overview of the palmprint de-identification and recognition pipeline is illustrated in Fig. \ref{fig:pr_de-id}. As shown on the left side, the goal of the de-identified image is 1) to increase the feature distance from the original in the identity space, thereby denying recognition; 2) preserving the image’s quality and usability as closely as possible to the original.

This paper presents an intuitive and effective framework for palmprint de-identification, leveraging a general pre-trained inpainting diffusion model, Paint by Example \cite{yang2023paint}, as the core component. The model is employed to regenerate palmprint regions in a manner that conceals identity-related information. However, since this diffusion model is not tailored specifically for palmprints, its direct application can yield unstable or unpredictable results. Fine-tuning or retraining such models to handle palmprint data typically requires access to large-scale and diverse datasets, which may not always be feasible. To address this, we introduce a training-free and optimization-free enhancement approach by incorporating two key strategies: \textit{semantic-guided embedding fusion} and \textit{prior interpolation}. By manipulating the guidance and constraint of the generating process, these mechanisms can improve the stability and controllability of the generation process without additional training. Furthermore, the inherent diversity of diffusion models enables the generation of multiple distinct, de-identified outputs from a single input, thereby breaking the deterministic one-to-one mapping and mitigating the risk of reverse inference.

Our framework operates as a black-box solution, requiring only a palmprint image as input and making no assumptions about underlying recognition models or auxiliary information. Besides, there is a lack of an informative and intuitive metric for de-identification performance in previous works. To quantitatively assess the effectiveness of de-identification, we propose a new evaluation metric: the \textit{de-identification ratio}, which measures the dissimilarity of identity features by a multi-dimensional and comprehensive index.

Extensive experiments conducted across several palmprint datasets and recognition systems, including both controlled and challenging scenarios, demonstrate that the proposed method offers strong de-identification capabilities. It effectively preserves non-identity-related information and usability. Additionally, our method exhibits high output diversity, producing distinctly different de-identified versions from the same palmprint input. Comparative analyses against conventional anonymization techniques, such as masking, blurring, and pixelating, highlight the superiority of our approach in striking a balance between identity concealment and image usability.

The primary contributions of this work are summarized as follows:

1.	To the best of our knowledge, this is the first study specifically designed to target the task of palmprint de-identification.

2.	We propose an intuitive and effective framework that utilizes a general pre-trained inpainting diffusion model for palmprint de-identification. To enhance stability and controllability in the generation process, we introduce \textit{semantic-guided embedding fusion} and \textit{prior interpolation} strategies. Notably, the framework operates in a training-free and optimization-free manner, requiring only a palmprint image as input.

3.	We introduce a novel metric, the \textit{de-identification ratio}, designed to more comprehensively assess de-identification effectiveness at the feature level, capturing both identity suppression and information preservation.

4.	Experimental results on diverse palmprint datasets and recognition methods demonstrate the effectiveness, generality, and robustness of our approach. The generated outputs exhibit high diversity, which helps mitigate the risk of inverse inference.

The rest of the paper is organized as follows: Section \ref{sec:2} reviews related work, Section \ref{sec:3} details the diffusion model and proposed framework, Section \ref{sec:4} presents experimental evaluations, Section \ref{sec:5} discusses key design choices, benefits, and flexibility of the proposed framework, and Section \ref{sec:6} concludes with a summary and future research directions.

\section{Related Works}\label{sec:2}
In this section, we briefly review related works in palmprint recognition and biometrics de-identification.

\subsection{Palmprint Recognition}\label{sec:2.1}
Palmprint recognition is a well-established biometric technique that typically involves several key stages: Region of Interest (ROI) localization, feature template extraction, template matching, and final decision-making. As illustrated in the right section of Fig. \ref{fig:pr_de-id}, the ROI localization step isolates the texture-rich region of the palm, which serves as the input for subsequent identity recognition processes. Feature templates are then extracted from the ROI and either stored in a database during the enrollment phase or compared against stored templates during the identification phase using matching algorithms. The final decision is made based on the similarity score between templates.

Most existing research focuses on improving recognition performance using the palmprint ROI. For example, hand-crafted methods such as \cite{yang2023multi} utilize Gabor filters to extract multiple directional features, while deep learning approaches \cite{yang2023comprehensive,yang2025multi} employ convolutional neural networks to learn discriminative texture features. Other works aim to enhance ROI extraction techniques under varying conditions. These include keypoint-based localization \cite{liang2023pklnet}, which predicts valley points between fingers to define the ROI; complete ROI extraction methods \cite{su2024complete} that encompass the full palmprint area; and techniques designed for ROI extraction in unconstrained environments \cite{matkowski2019palmprint}. A separate line of research moves beyond the ROI, using the entire hand image for recognition, as seen in full-hand methods \cite{matkowski2023improving} that align the hand using anatomical landmarks such as finger and palm regions.

In terms of privacy protection, previous studies have primarily focused on securing feature templates. For instance, the work in \cite{liu2024palmsecmatch} leverages the stochastic nature of biometric data to protect palmprint templates, while \cite{yang2024dual} proposes a dual-level cancelable palmprint verification framework to enhance security granularity. However, a growing body of research has begun to explore palmprint-level attacks. These include reconstruction attacks \cite{yan2024toward} that regenerate palmprint textures from templates, even under data-limited conditions; backdoor attacks \cite{wang2025gan} that utilize GANs to inject hidden triggers into palmprint images; and ROI embedding attacks \cite{yan2024realistic} that insert malicious ROIs into benign hand images, enabling covert compromises in realistic scenarios.

In conclusion, while palmprint recognition fundamentally relies on the texture information within the ROI, the increasing sophistication of image-level attacks has highlighted a critical gap in privacy protection. Existing approaches largely overlook the image-level privacy threat, making current recognition systems vulnerable to attacks that fake or steal palm textures—potentially undermining the integrity of palmprint-based biometric authentication.

\subsection{Biometrics De-identification}\label{sec:2.2}
Biometric de-identification, also known as anonymization, refers to the process of obscuring identity-specific information in biometric data while preserving non-identifying attributes to maintain usability \cite{shopon2021biometric}. This typically involves modifying or replacing personal identifiers in a way that conceals sensitive information from public access. The de-identification pipeline, illustrated in Fig. \ref{fig:pr_de-id}, using palmprint as an example, aims to increase the disparity between the original and de-identified features to prevent successful recognition while simultaneously preserving the visual quality and usability of the de-identified sample.

In the context of face de-identification, early approaches, such as \cite{gross2006model}, employed model-based techniques that extracted and blended appearance features from different individuals to produce de-identified images. More advanced frameworks, such as the four-stage system proposed in \cite{li2019anonymousnet}, introduce a combination of attribute obfuscation, generative reconstruction, and adversarial perturbation to achieve de-identification. Subsequently, GAN-based and adversarial-based methods gained popularity, including U-Net-based GANs that overlay external facial features onto the original face \cite{lin2021fpgan}, and adversarial perturbation techniques that distort images to evade recognition \cite{ghafourian2023toward}. Recently, diffusion models have emerged as powerful tools for face de-identification, either through model training tailored to the task \cite{he2024diff} or by using guided text prompts to manipulate facial identity \cite{sun2024diffam}.

For other biometric modalities, palm vein privacy protection has been explored in \cite{abd2020data} using feature-level encryption, while fingerprint de-identification has been attempted through high-level semantic noise injection \cite{li2024hierarchical}. Additionally, \cite{panigrahy2009privacy} proposed an image-level encryption scheme for face, palmprint, and signature data. However, these methods typically produce images that lack visual realism due to severe distortions, limiting their practical applicability.

Overall, existing methods either fail to maintain the natural appearance of the de-identified data or depend on complex procedures and additional inputs. Notably, no prior work has addressed palmprint de-identification in a way that concurrently ensures both visual fidelity and robust privacy protection.

\section{Methodology}\label{sec:3}
\subsection{Preliminary}\label{sec:3.1}
\subsubsection{Stable Diffusion}\label{sec:3.1.1}
Stable Diffusion (SD) \cite{ho2020denoising,rombach2022high} is a cutting-edge latent diffusion model that enables high-fidelity image generation and manipulation from either textual or visual prompts. It operates via a two-stage denoising framework. In the forward process, Gaussian noise is progressively injected into the image until it transforms into pure noise. Conversely, in the reverse denoising process, a U-Net architecture trained on noisy latent representations reconstructs the image by gradually removing noise conditioned on semantic-guided embeddings, such as textual or visual embeddings from models like CLIP \cite{radford2021learning}.

A key aspect of SD is its use of a Variational Autoencoder (VAE) \cite{kingma2013auto} to map images into a compact latent space, significantly reducing computational overhead without compromising critical visual information. This latent-space modeling allows SD to efficiently learn the underlying data distribution, resulting in high-quality image synthesis with fine-grained control over generated content.

\subsubsection{Paint by Example}\label{sec:3.1.2}
Paint by Example \cite{yang2023paint} is an exemplar-based image inpainting diffusion model built upon the Stable Diffusion (SD) framework. It is trained in a self-supervised manner to reconstruct masked regions of an image by leveraging both a background image and an exemplar image from a same image as conditioning signals. Specifically, the model takes a masked background image and an exemplar image as inputs, leveraging contextual cues from the background and semantic-guided embeddings from the exemplar to complete the missing region. In addition, the contextual cues are encoded into latent maps through the VAE encoder. Importantly, the exemplar image undergoes a semantic compression process through a bottleneck architecture composed of CLIP and a multilayer perceptron (MLP), ensuring that only high-level semantic features, not detailed identity-specific information, are retained. This prevents direct leakage of exemplar details into the inpainting result.

Additionally, the model's reliance on stochastic initialization via random noise allows it to generate diverse outputs even under identical conditions, a property highly desirable for de-identification tasks where output variability is essential. Given these characteristics, i.e., semantic-level control, detail suppression, context preservation, and generation diversity, Paint by Example is well-suited for the palmprint de-identification problem, where the goal is to obscure identity while maintaining a natural appearance and usability.

\subsection{Palmprint De-identification}\label{sec:3.2}
\subsubsection{Overall Pipeline of Our Method}\label{sec:3.2.1}
\begin{figure*}[!t]
	\centering
	\includegraphics[width=1\textwidth]{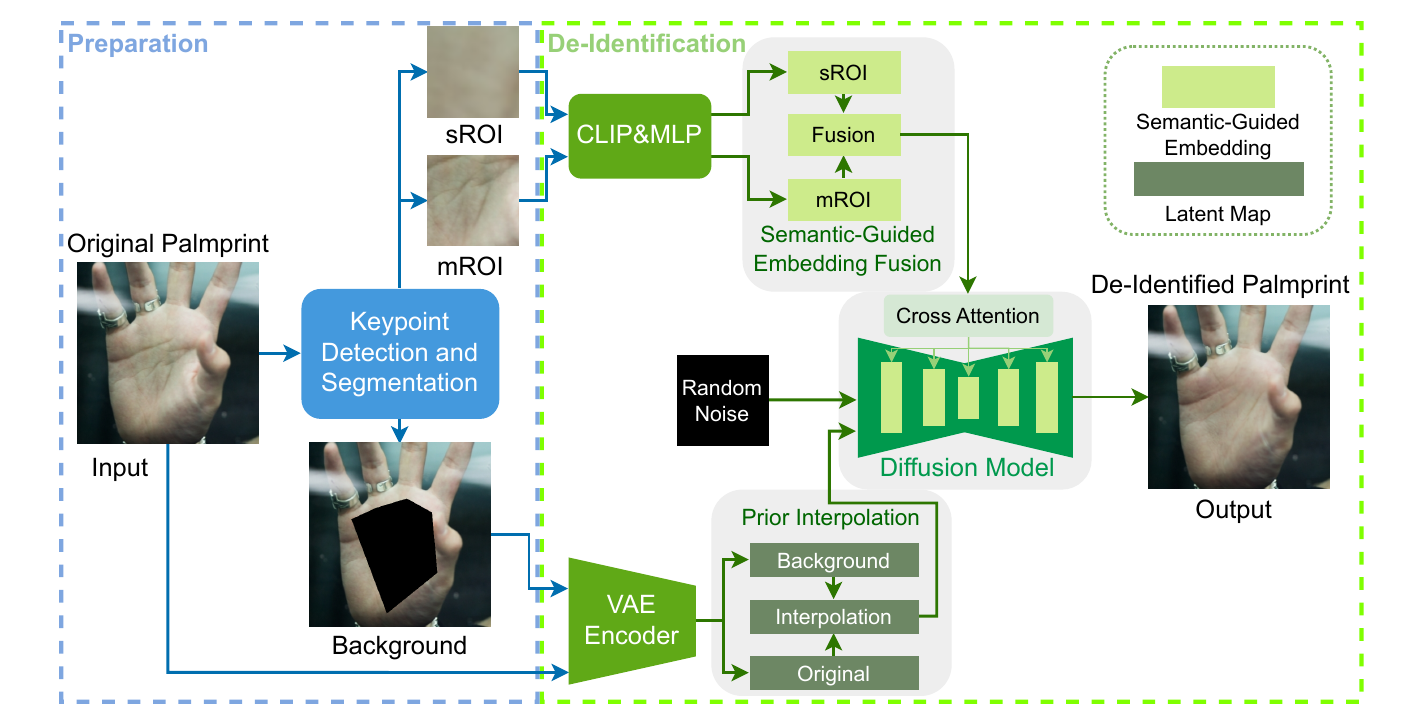}
	\caption{Overall pipeline of our proposed method. 1) The preparation step will provide the essential exemplar ROI images and a background image with a mask. The sROI and mROI represent the small ROI (sROI) and a medium ROI (mROI), respectively. 2) The de-identification step will generate a new palmprint for the masked area with the fusion semantic guidance and interpolation latent context.}
	\label{fig:pipeline}
\end{figure*}

Based on the Paint by Example model \cite{yang2023paint}, we propose a training-free and optimizing-free palmprint de-identification pipeline, as shown in Fig. \ref{fig:pipeline}. The pipeline is composed of two steps: preparation and de-identification.

In the preparation step, all required components are extracted from a single hand image using a keypoint detection model and a segmentation model (Section \ref{sec:3.2.2}). Two regions of interest, a small ROI (sROI) and a medium ROI (mROI), are isolated to serve as exemplar inputs to extract semantic-guided embedding (SGE). Simultaneously, a masked background image is obtained for constraining the inpainting process. Both the original palmprint and the masked background are essential inputs for the subsequent de-identification step.

In the de-identification step, key components, including CLIP and MLP, the VAE Encoder, and the diffusion model, are inherited from the Paint by Example framework. The CLIP\&MLP module functions as a semantic bottleneck, extracting highly compressed SGE from the sROI and mROI. Both SGEs are then fused into a unified representation via a \textit{semantic-guided embedding fusion} strategy (Section \ref{sec:3.2.3}), capturing richer and complementary identity-independent features.

Meanwhile, the VAE encoder transforms both the original and background images into their respective latent maps. A \textit{prior interpolation} strategy is then employed to compute a new latent map that lies between the original and background latent maps. This interpolated latent map acts as a controllable modulator, enabling a balance between effective de-identification and the preservation of image quality and usability (refer to Section \ref{sec:3.2.4}).

Finally, the diffusion model synthesizes the masked region by integrating the fusion SGE and conditioning on the interpolated latent (refer to Section \ref{sec:3.2.5}). The result is a high-quality, realistic palmprint image with de-identified textures that maintain usability while obscuring identity traits. Each component of this pipeline is elaborated in the subsequent sections.

\subsubsection{Keypoint Detection and Segmentation}\label{sec:3.2.2}
\begin{figure}[!h]
	\centering
	\includegraphics[width=1\textwidth]{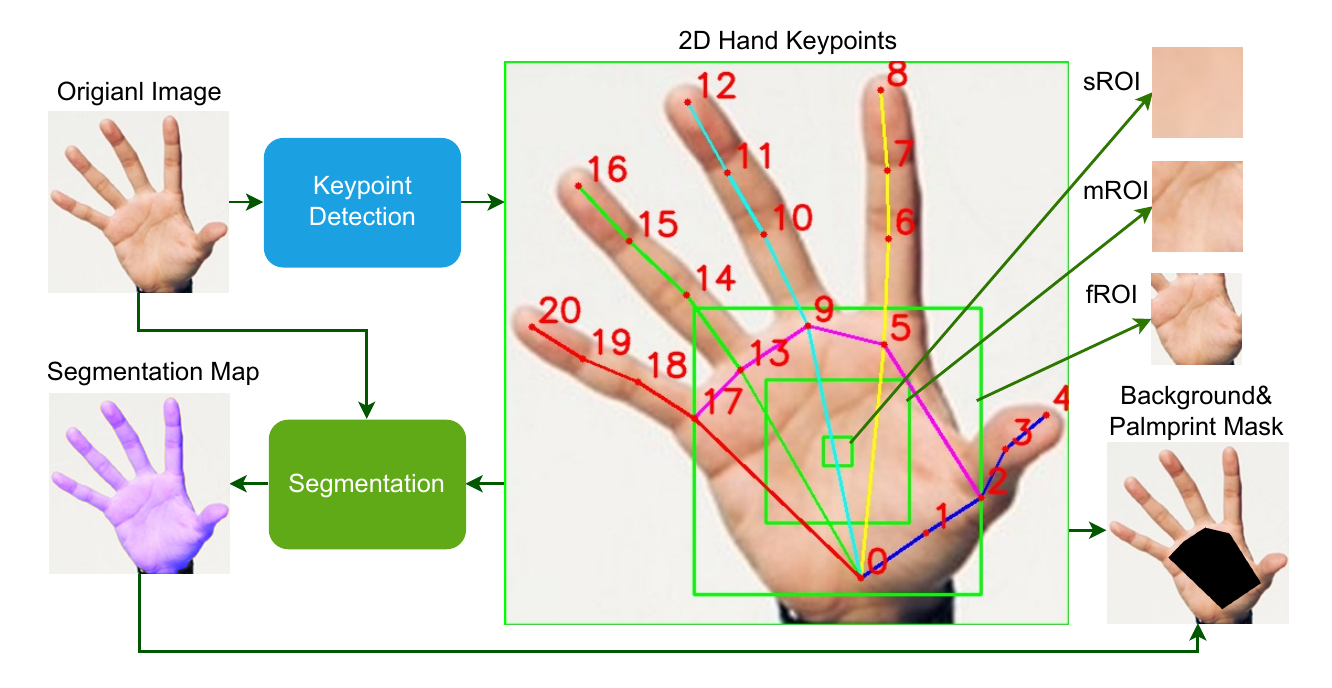}
	\caption{Pipeline of hand keypoint detection and segmentation. The keypoints are applied to extract the full-scale ROI (fROI), the medium ROI (mROI), the small ROI (sROI), and the palmprint mask. The hand segmentation map is adopted to refine the palmprint mask and preserve the background.}
	\label{fig:kp_seg}
\end{figure}

The pipeline of the Hand Keypoint Detection and Segmentation module is shown in Fig. \ref{fig:kp_seg}. A pre-trained wild hand keypoint detection model \cite{potamias2024wilor} is applied to extract the hand 2D keypoints, and the SAM2 \cite{ravi2024sam} is adopted for hand segmentation.

Firstly, 21 hand keypoints are extracted from the original image using the keypoint detection model. Among these, seven keypoints i.e. 0, 1, 2, 5, 9, 13, and 17, as shown in Fig. \ref{fig:kp_seg}, are used to define an encompassed region corresponding to the texture-rich palm area. This region serves as the masked area for inpainting, aligning with conventional palmprint ROI localization. Simultaneously, preserving the unmasked palmprint region is critical, as the inpainting model requires sufficient contextual palmprint information in the background image for realistic synthesis.

Subsequently, a full-scale ROI (fROI) is defined by computing a minimum bounding square enclosing the seven selected keypoints. However, directly using the fROI as a semantic exemplar can introduce semantic noise and structural distortions due to its relatively broad and unrefined content. To address this, the fROI is rescaled to 50\% and 10\% of its original size to obtain two exemplar regions: the medium ROI (mROI) and the small ROI (sROI). These ROIs are designed as input to extract clean, complementary SGE during de-identification. Before input into the de-identification module, both ROIs are resized to a standard resolution of 128×128 to ensure uniformity.

Finally, all 21 keypoints are employed to guide SAM2 in generating a precise hand segmentation map. This map serves two purposes: first, to refine the masked area by removing regions erroneously included due to the bounding region around the seven keypoints; and second, to enable seamless integration of the de-identified hand back into the original image, thereby preserving the background with high fidelity.

\subsubsection{Semantic-guided Embedding Fusion}\label{sec:3.2.3}
In the original Paint by Example framework, only a single image is used as the exemplar. Semantic-guided embedding (SGE) compresses the exemplar image through a narrow bottleneck before conditioning the diffusion U-Net, thereby reducing identity leakage—such as fine-grained palmprint textures—while retaining higher-level semantic cues, such as overall palmprint appearance. Nevertheless, this mechanism is not specifically optimized for palmprint generation and may yield unpredictable or suboptimal results, often caused by inaccurate semantic representation or interference from semantic noise. To address this limitation, we introduce a simple yet effective SGE fusion strategy that integrates SGE from exemplar ROIs at multiple scales. This multi-scale fusion enhances both the reliability and stability of the guidance signal. The fusion process is formalized as follows:

\begin{equation}
	\label{eq:1}
	\overline{g} = \frac{1}{n} \sum\limits_{i=1}^n g_i,
\end{equation}

where $n$ indicates the total number of SGE, and $g$ and $\overline{g}$ represent the single SGE and the final fusion SGE, respectively.

Since SGE functions as a semantic representation, our fusion strategy effectively seeks a balanced midpoint among these representations. By fusing SGE across multiple scales, the proposed approach mitigates noise or distortions originating from a single exemplar and reinforces semantic consistency. As a result, the fusion process enhances the stability and quality of the generated outputs, producing more consistent and desirable results.

\subsubsection{Prior Interpolation}\label{sec:3.2.4}
In addition to exemplar semantic guidance, the inpainting results are also influenced by the latent representation of the background. When the background contains undesirable context, it can adversely affect the final output. A natural solution is to incorporate prior knowledge from the original image to guide the generation. By incorporating prior structural information from the original image, the generation process can be stabilized and constrained by the given structural context. However, this must be done carefully; excessive prior information risks compromising the de-identification objective. To address this trade-off, we propose a \textit{prior interpolation} strategy that allows fine-grained control over the balance between visual quality and identity obfuscation. The \textit{prior interpolation} mechanism is defined as follows:

\begin{equation}
	\label{eq:2}
	z_{\text{in}} = \alpha z_{\text{o}} + (1 - \alpha) z_{\text{bg}},
\end{equation}

where $z_{\text{bg}}$, $z_{\text{o}}$, and $z_{\text{in}}$, denote background, original, and interpolation latent map, respectively. Here, $\alpha$ serves as the interpolation factor, ranging from 0 to 1. A higher $\alpha$ shifts the interpolated latent representation closer to the original image’s latent space, while pulling it further away from the background latent. This provides a controllable mechanism to modulate the influence of original content versus background context during inpainting. 

Here, larger $\alpha$ values favor usability and perceptual fidelity, whereas smaller $\alpha$ values enhance identity suppression at the potential cost of realism. Intuitively, minimizing identity leakage require a small $\alpha$ for prior context guidance. In our experiments, we set $\alpha = 0.1$ as a balanced point, achieving a desirable trade-off between identity similarity, perceptual fidelity, and usability across datasets.

\subsubsection{De-identified Palmprint Synthesis}\label{sec:3.2.5}

The generation of de-identified palmprints in the diffusion model follows the same principle as Paint by Example \cite{yang2023paint}. To synthesize de-identified palmprints, the fused SGE is injected into each block of the U-Net within the diffusion model using a cross-attention mechanism \cite{vaswani2017attention}, progressively steering the denoising and hence generation process. Simultaneously, the interpolated latent representation and its corresponding mask are fed into the U-Net, along with random noise. During generation, the unmasked regions of the latent remain unchanged, preserving their original structure, while the masked regions are regenerated from a random noise. This regeneration is both guided by the fused SGE and constrained by the context of the surrounding unmasked areas, ensuring semantic consistency and structural realism.

\subsection{Evaluation Metrics}\label{sec:3.3}
The evaluation of de-identification methods is typically approached from three complementary perspectives: effectiveness of de-identification, image quality, and preservation of usability.

To assess image quality, we employ a set of five diverse and widely accepted metrics: Structural Similarity (SSIM), Multi-Scale SSIM (MS-SSIM), Peak Signal-to-Noise Ratio (PSNR), Learned Perceptual Image Patch Similarity (LPIPS) \cite{zhang2018unreasonable}, and Fréchet Inception Distance (FID) \cite{heusel2017gans}, to ensure a thorough and multifaceted evaluation of visual fidelity and perceptual similarity.

Usability preservation is measured by quantifying the differences between original and de-identified images across four downstream tasks: hand detection, keypoint detection, ROI localization, and hand segmentation. This reflects the extent to which functional utility is retained despite the removal of identity-specific cues.

The de-identification performance is evaluated under both verification and recognition protocols. In recognition tasks, a lower classification accuracy indicates stronger de-identification, as it implies reduced identity leakage. Conversely, in verification scenarios, a higher matching distance signifies better anonymization, as it reflects greater dissimilarity from the original biometric. Verification decisions rely on a threshold, which we determine based on the Equal Error Rate (EER)—a commonly used operating point that balances false acceptances and rejections. If a de-identified sample fails verification under this threshold, it is treated as a rejection, enabling us to compute the Rejection Rate (RR) as an indicator of performance.

However, RR is inherently threshold-sensitive and fails to capture the overall distributional shift introduced by de-identification. To address this limitation, we propose a more robust and informative metric: the \textit{De-identification Ratio} (DIR), grounded in the decidability index $d'$ \cite{daugman2003importance}. The $d'$ index quantifies the separability between two statistical distributions $D1$ and $D2$, and is computed as follows:

\begin{equation}
	\label{eq:3}
	d'(D_1,D_2) = \frac{\left| \mu_1 - \mu_2 \right|}{\sqrt{\frac{\sigma_1^2 + \sigma_2^2}{2}}},
\end{equation}

where $\mu_n$ and $\sigma_n$ denote the mean and standard deviation of distribution $n$ $(n \in {1,2})$, respectively.

The DIR is designed to evaluate the distribution of de-identified match scores by referencing both genuine and imposter matching distributions, which serves as an index indicating whether the de-identified matches are closer to the genuine or imposter distribution. Ideally, de-identified matches should lie closer to the imposter distribution and farther from the genuine distribution. In particular, DIR evaluates de-identification performance from the matching distribution perspective, avoiding sensitivity to threshold selection or the influence of abnormal matches. This enables a comprehensive, multidimensional assessment of de-identification performance. Formally, the DIR can be defined as follows:

\begin{equation}
	\label{eq:4}
	\text{DIR} = \frac{d'(D_g,D_d)}{d'(D_g,D_i)} = \frac{\mu_g - \mu_d}{\mu_g - \mu_i} \times \frac{\sqrt{\sigma_g^2 + \sigma_i^2}}{\sqrt{\sigma_g^2 + \sigma_d^2}} \times 100\%,
\end{equation}

where the subscripts $g$, $i$, and $d$, represent the genuine, imposter, and de-identification distribution.

This metric captures the relative difference between the genuine–de-identified and genuine–imposter distributions, effectively quantifying how well the de-identified data mimics the statistical behavior of imposters. A DIR value of 100\% represents the ideal outcome, indicating that the de-identified samples are indistinguishable from imposter samples. In this case, the recognition system treats the de-identified sample as belonging to a different individual, reflecting successful anonymization. In contrast, a DIR near or below 0\% suggests that the de-identified samples remain highly similar to the originals at the feature level, revealing that the de-identification strategy fails under the given recognition model.

For interpretability, we define qualitative bands of identity suppression effectiveness: DIR $>$ 80\% denotes high suppression effectiveness, values in the range of 60–80\% indicate moderate suppression effectiveness, and DIR $<$ 60\% reflects limited suppression effectiveness. Generally, the closer the DIR is to 100\%, the more effective the de-identification. However, values exceeding 100\% should be avoided, as they indicate that the de-identified features deviate even more than typical imposters, possibly falling outside the expected distribution of normal samples and introducing recognition instability.

Importantly, DIR scores are dependent on the recognition model. A single set of de-identified samples may yield varying DIR values across different systems, as each method captures and prioritizes distinct aspects of palmprint texture.

\section{Experiments}\label{sec:4}
\textbf{Palmprint recognition:} Three methods are selected to evaluate the de-identification performance, which are the MTCC \cite{yang2023multi} (hand-crafted-based), CCNet \cite{yang2023comprehensive} (deep-learning-based), and EEHNet \cite{matkowski2023improving} (full-hand-based) methods. PKLNet \cite{liang2023pklnet} is applied to extract all of the ROIs for palmprint recognition in our experiments, and the ROI’s size is 128×128 for all datasets. The genuine and imposter distribution is calculated on the entire dataset for each dataset.

\textbf{Datasets:} Five contactless hand datasets are included in our experiments, as detailed in Tab. \ref{tab:datasets}, and their visual presentation is shown in Fig. \ref{fig:exp_de-id}. 

\begin{table}[!h]
	\caption{Datasets introduction. The details of five contactless hand datasets.\label{tab:datasets}}
	\centering
	\begin{tabular}{c|cccc}
		\toprule
		\textbf{Name}   & \textbf{Identities} & \textbf{Samples} & \textbf{Environment} & \textbf{Resolution} \\
		\midrule
		IITD  \cite{iitd} & 460        & 2,601   & Controlled  & 1600×1200  \\
		PolyU \cite{polyu} & 177        & 1,770   & Controlled  & 640×480    \\
		REST \cite{charfi2016local}  & 358        & 1,948   & Controlled  & 2480×1536  \\
		Zhou \cite{zhou2019key}  & 166        & 1,295   & Wild        & 1080×1920  \\
		NTU-PI \cite{matkowski2019palmprint} & 2,035      & 7,881   & Wild        & 227×227  \\
		\bottomrule
	\end{tabular}
\end{table}

\textbf{Implementation details:} Experiments are performed using a single RTX 4090 GPU, with each identified sample generated within approximately 2 to 3 seconds, primarily influenced by the computational cost of the diffusion process. The input image can be of arbitrary size, as Wilor \cite{potamias2024wilor} is employed to detect the hand region and seamlessly reinsert it into the original context. To ensure consistency, a fixed random seed is used in the diffusion model, controlling the generation of stochastic noise during sampling.

\subsection{De-identification Performance}\label{sec:4.1}
We selected two distinct palmprint recognition approaches for evaluation: MTCC, a hand-crafted feature-based method \cite{yang2023multi}, and CCNet, a deep learning-based model \cite{yang2023comprehensive}. Although our primary focus lies on palmprint regions, we also extended our analysis to include full-hand recognition by testing EEHNet \cite{matkowski2023improving}. Tab. \ref{tab:de-id} presents the comparative performance across various datasets and recognition methods. In the table, the symbols ↑ and ↓ indicate whether a higher or lower value is preferable, respectively.

\begin{table}[!t]
	\caption{De-identification performance (\%). The symbols ↑ (↓) indicate that the higher (lower) value is better.\label{tab:de-id}}
	\resizebox{1\linewidth}{!}{
	\centering
	\begin{tabular}{c|ccc|ccc|ccc}
		\toprule
		\multirow{2}{*}{\textbf{Dataset}} & \multicolumn{3}{c|}{\textbf{MTCC}} & \multicolumn{3}{c|}{\textbf{CCNet}} & \multicolumn{3}{c}{\textbf{EEHNet}} \\
		& \textbf{RR↑}    & \textbf{DIR↑}   & \textbf{Acc.↓}  & \textbf{RR↑}    & \textbf{DIR↑}   & \textbf{Acc.↓}  & \textbf{RR↑}    & \textbf{DIR↑}   & \textbf{Acc.↓}  \\
		\midrule
		IITD                     & 95.42  & 94.79  & 10.35  & 96.42 & 82.55 & 10.12  & 41.77   & 66.14   & 49.77  \\
		PolyU                    & 93.73  & 97.47  & 6.67   & 62.66 & 81.55  & 13.05   & 38.53   & 68.43   & 69.72  \\
		REST                     & 49.85  & 85.63  & 6.47   & 51.08   & 84.40  & 9.19   & 6.31    & 27.89   & 56.88  \\
		Zhou                     & 79.61  & 91.44  & 10.91  & 31.76   & 75.49  & 27.17  & 34.31   & 68.86   & 72.21  \\
		NTU-PI                   & -      & -      & -      & -       & -      & -      & 5.54    & 0.26    & 60.97 \\
		\bottomrule
	\end{tabular}
}
\end{table}

As summarized in Tab. \ref{tab:de-id}, our de-identification method demonstrates strong performance across various recognition models and datasets. Specifically, in the hand-crafted MTCC method, our approach achieves a minimum of 85.63\% DIR and a maximum accuracy of 10.91\%. In contrast, for CCNet, the DIR reaches 75.49\% at least, while accuracy peaks at 27.17\% at most. Notably, on the REST dataset, we observe a favorable DIR of 84.40\% and a peak accuracy of just 9.19\%, despite a low RR of 51.08\%. This discrepancy highlights that DIR offers a more reliable measure of de-identification efficacy compared to RR.

Given that the NTU-PI dataset lacks suitability for palmprint ROI-based recognition, it was evaluated exclusively with the full-hand recognition method, EEHNet. Interestingly, even though our method targets palmprint-specific de-identification, EEHNet still delivers promising verification performance on the IITD, PolyU, and Zhou datasets. NTU-PI, however, shows the weakest performance, likely due to its limited reliance on palmprint features for identity verification.

Furthermore, the observed de-identification results between MTCC and CCNet are not directly aligned, despite both being evaluated on the identical palmprint ROI. This divergence stems from the fact that each method emphasizes different discriminative features within the palmprint texture. Our proposed approach, functioning as a model-agnostic (black-box) solution, does not assume prior knowledge of which features are most influential for each recognition method. Nonetheless, it consistently achieves effective de-identification performance across diverse models and datasets.

\subsection{Quality of De-Identified Palmprints}\label{sec:4.2}
\begin{figure}[!t]
	\centering
	\includegraphics[width=1\linewidth]{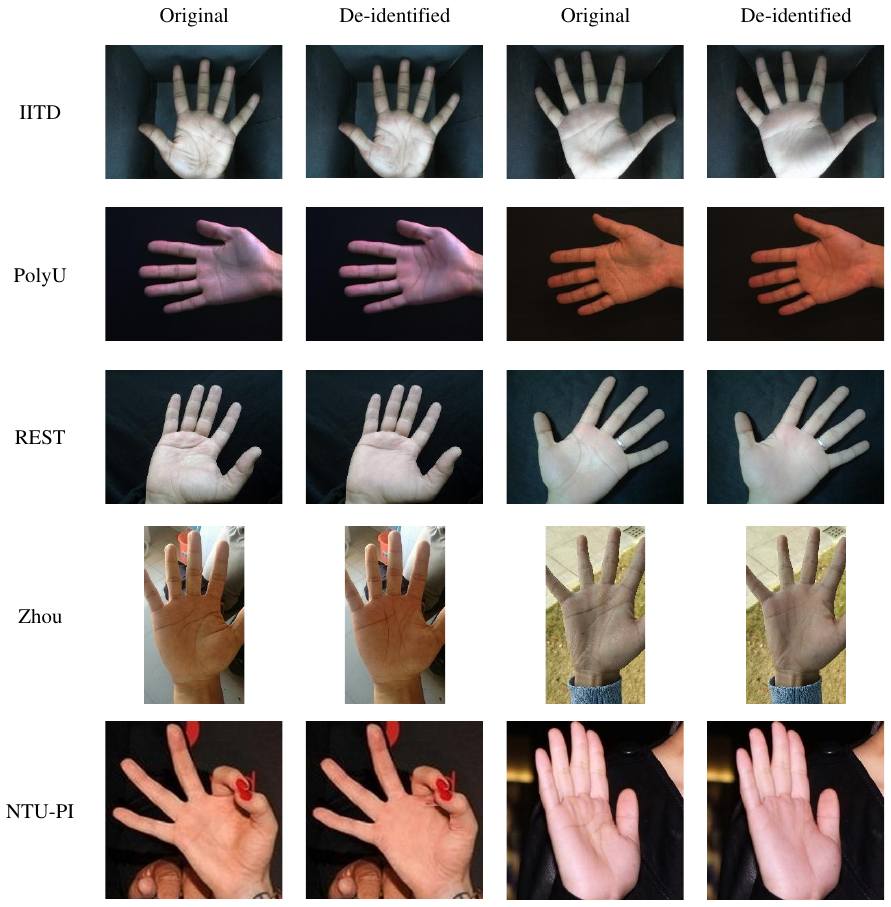}
	\caption{Visual presentation of palmprint de-identification. This figure shows two pairs of original-de-identified images for the five datasets.}
	\label{fig:exp_de-id}
\end{figure}

Fig. \ref{fig:exp_de-id} illustrates the visual outcomes of our de-identification method across various palmprint datasets. As observed, the generated images retain high visual fidelity, making it difficult, even upon close inspection, to distinguish them from their original counterparts without explicit cues. This highlights the effectiveness of our approach in preserving perceptual realism.

To quantitatively assess the quality of the de-identified images, we evaluated their similarity to the original images using five established metrics: SSIM, MS-SSIM, PSNR (where higher values indicate better quality), LPIPS and FID (where lower values are preferred). The results, presented in Tab. \ref{tab:quality}, confirm that our method introduces minimal distortion while preserving the original style and structural consistency. This balance between visual realism and subtle identity modification underscores the strength of our de-identification strategy.

\begin{table*}[!t]
	\caption{De-identified image quality. The symbols ↑ (↓) indicate that the higher (lower) value is better.\label{tab:quality}}
	\resizebox{1\linewidth}{!}{
	\centering
	\begin{tabular}{c|ccccc}
		\toprule
		\textbf{Dataset} & \textbf{SSIM↑}         & \textbf{MS-SSIM↑}      & \textbf{PSNR↑}          & \textbf{LPIPS↓}        & \textbf{FID↓}   \\
		\midrule
		IITD    & 0.9633±0.0104 & 0.9628±0.0106 & 32.656±2.2638  & 0.1064±0.0252 & 6.8021 \\
		PolyU   & 0.9439±0.0099 & 0.9639±0.0082 & 34.1753±1.5226 & 0.1156±0.0134 & 7.7344 \\
		REST    & 0.9707±0.0066 & 0.9695±0.007  & 35.3343±1.7182 & 0.0753±0.0145 & 4.9845 \\
		Zhou    & 0.9518±0.0244 & 0.9586±0.0147 & 33.6194±2.6124 & 0.1148±0.0278 & 8.4354 \\
		NTU-PI  & 0.9613±0.0191 & 0.9427±0.0237 & 31.3071±3.1251 & 0.096±0.0213  & 4.5092 \\
		\bottomrule
	\end{tabular}
}
\end{table*}

\subsection{Usability after De-identification}\label{sec:4.3}
Usability plays a critical role in evaluating de-identification methods. An ideal approach should effectively remove identity-related features while preserving the image's usability for other downstream tasks. In our study, we assess five key usability aspects: hand detection, 2D and 3D keypoint detection provided by Wilor \cite{potamias2024wilor}, as well as ROI localization and hand segmentation provided by PKLNet \cite{liang2023pklnet}.

Usability is quantified by computing the mean absolute error between the outputs generated from the original and de-identified images. To facilitate consistent comparison, all differences are normalized to a [0–100]\% scale, where lower values indicate better preservation of usability.

As shown in Tab. \ref{tab:usability}, our method maintains a usability difference of less than 1\% in most scenarios, indicating that the de-identified images remain highly suitable for various non-identification tasks. This demonstrates that our framework achieves a desirable balance between identity removal and functional integrity.

\begin{table}[!t]
	\caption{Usability differences of de-identification ↓ (\%). The range of the value is [0-100]\%, and the lower value is the better.\label{tab:usability}}
	\resizebox{1\linewidth}{!}{
	\centering
	\begin{tabular}{c|ccccc}
		\toprule
		\textbf{Dataset} & \makecell{\textbf{Hand} \\ \textbf{Detection}} & \makecell{\textbf{2D Keypoint} \\ \textbf{Detection}} & \makecell{\textbf{3D Keypoint} \\ \textbf{Detection}} & \makecell{\textbf{ROI} \\ \textbf{Localization}} & \makecell{\textbf{Hand} \\ \textbf{Segmentation}} \\
		\midrule
		IITD    & 0.235±0.223 & 0.258±0.096 & 0.186±0.108 & 0.146±0.255 & 0.182±0.151 \\
		PolyU   & 0.202±0.226 & 0.256±0.086 & 0.179±0.098 & 0.200±0.557   & 0.496±0.645 \\
		REST    & 0.320±0.389  & 0.263±0.140  & 0.176±0.112 & 0.128±0.347 & 0.267±0.213 \\
		Zhou    & 0.392±0.536 & 0.390±0.875  & 0.224±0.592 & 0.312±0.956 & 0.758±0.965 \\
		NTU-PI  & 0.675±2.041 & 0.751±3.363 & 0.283±0.828 & 4.457±4.501 & 5.148±5.026 \\
		\bottomrule
	\end{tabular}
}
\end{table}

\subsection{Diversity of De-identification}\label{sec:4.4}
Diversity is an essential characteristic for effective de-identification, as it mitigates the risk of inversion attacks and ensures unlinkability across multiple de-identified instances derived from the same source \cite{meden2021privacy}. A robust de-identification method should be capable of generating multiple distinctly different images, each significantly distant within the same identity class, from a single original image.

Our framework leverages the intrinsic diversity of the diffusion model to satisfy this requirement naturally. Diversity is driven by the stochastic nature of the initial noise input; different random noise seeds result in distinct generation outcomes. In the diversity experiment, we employ 10 distinct random seeds to generate 10 diverse de-identified samples per input image.

\begin{figure*}[!t]
	\centering
	\includegraphics[width=1\textwidth]{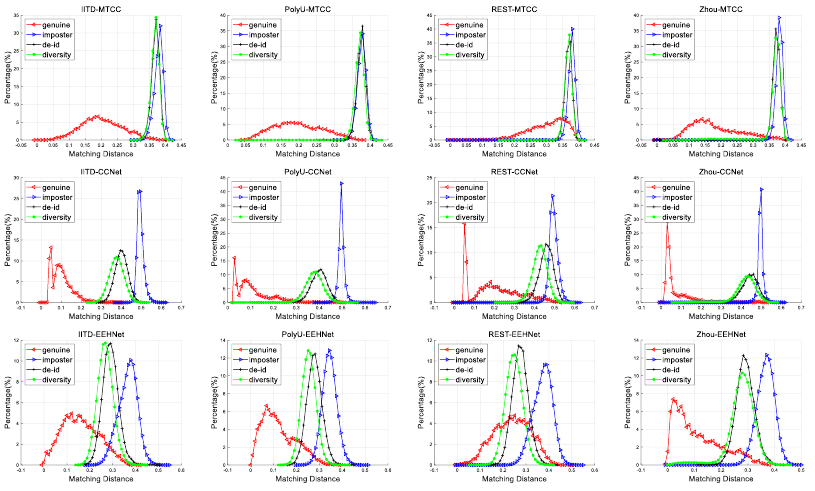}
	\caption{Distribution of de-identification diversity. The first, second, and third rows correspond to the MTCC, CCNet, and EEHNet methods, respectively, while the columns represent results across the IITD, PolyU, REST, and Zhou datasets. Each chart includes four curves: red (genuine matches), blue (imposters), black (original vs. de-identified), and green (inter-de-identified diversity).}
	\label{fig:exp_diversity}
\end{figure*}

To evaluate diversity, we analyze the distribution of matching distances, as visualized in Fig. \ref{fig:exp_diversity}. The first, second, and third rows correspond to the MTCC, CCNet, and EEHNet methods, respectively, while the columns represent results across the IITD, PolyU, REST, and Zhou datasets. The NTU-PI dataset is excluded due to suboptimal performance in this context. Each chart includes four curves: red (genuine matches), blue (imposters), black (original vs. de-identified), and green (inter-de-identified diversity). The black curve quantifies how effectively the de-identified images differ from their source, while the green curve reflects the spread among de-identified samples originating from the same image—serving as a direct measure of diversity.

\begin{figure}[!t]
	\centering
	\includegraphics[width=1\textwidth]{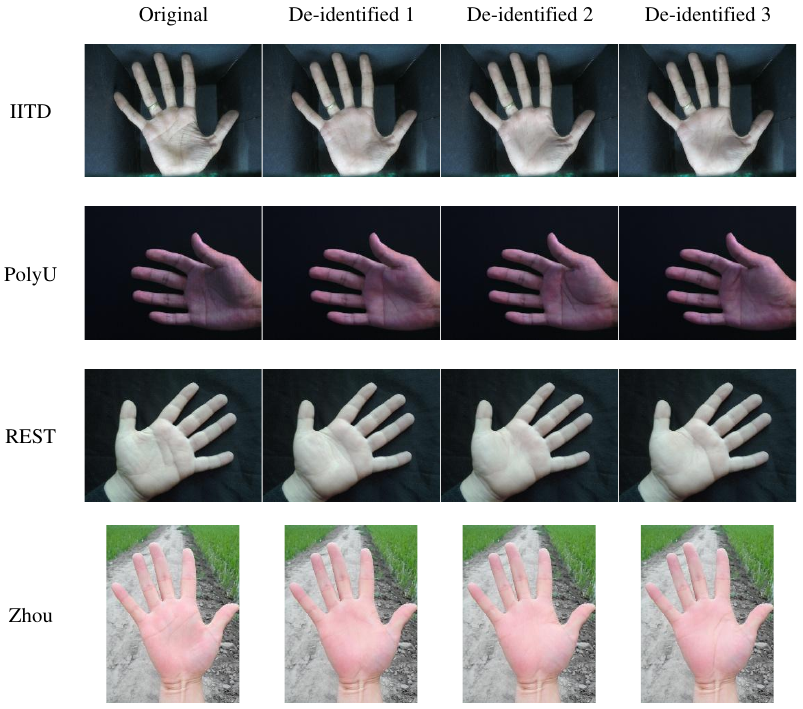}
	\caption{Visual presentation of de-identification diversity. The figure shows one original and three corresponding de-identified images from the original on four datasets.}
	\label{fig:exp_diversity_img}
\end{figure}

In the case of MTCC, the distributions for imposters, de-identified samples, and diversity are nearly overlapping—indicating that our method achieves both strong de-identification and high intra-class diversity simultaneously. 

For CCNet, while the diversity distribution is slightly closer to the genuine distribution than that of the de-identification distribution, it remains comparable to the imposter distribution, suggesting effective but slightly less varied sample generation. Remarkably, even in the full-hand scenario of EEHNet, our framework maintains strong performance, producing sufficiently diverse and untraceable outputs.

A qualitative illustration of this diversity is provided in Fig. \ref{fig:exp_diversity_img}. As evident, the de-identified palmprints generated from the same original image exhibit noticeably distinct textures, not only compared to the source palmprint but also among each other, clearly demonstrating the framework’s ability to produce diverse and unlinkable outputs.

\subsection{Ablation Study}\label{sec:4.5}
To comprehensively evaluate the effectiveness of our de-identification framework and explore the trade-offs between de-identification strength, image quality, and usability, we designed a series of targeted experiments. These experiments were divided into three major components: analyzing exemplar guidance, examining the role of latent map conditioning, and benchmarking against conventional anonymization techniques. All ablation studies were performed on the IITD dataset, with detailed descriptions of each experimental setting as follows:

\textbf{A.	ROI Scale Evaluation:}

This part investigates the effect of different exemplar ROI sizes—small (s), medium (m), and full (f)—on performance.

\textbf{B.	Semantic-guided Embedding Fusion Analysis:}

This experiment explores the fusion of multiple ROI scales, including combinations s+m, s+f, m+f, and s+m+f, where “+” denotes a fusion operation.

\textbf{C.	Prior Interpolation Factor ($\alpha$) Testing:}

Using the s+m fusion strategy, this setting varies the interpolation factor $\alpha$ among 0.1, 0.2, and 0.3 to analyze its effect on performance and image fidelity.

\textbf{D.	Comparison with Traditional Anonymization Methods:}

We benchmarked our method against standard techniques, including masking, blurring, and pixelation. 

\begin{figure}[!t]
	\centering
	\includegraphics[width=1\linewidth]{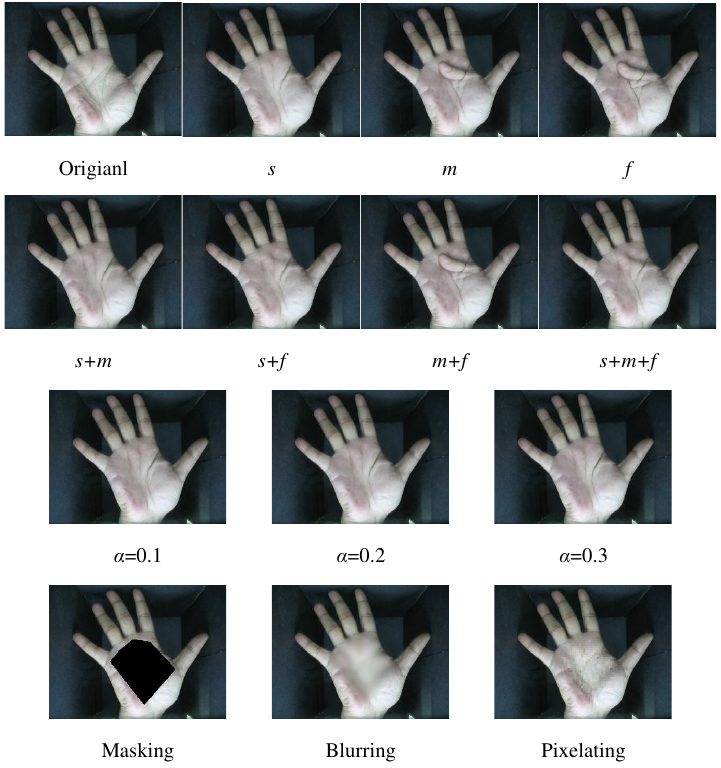}
	\caption{Visual presentation of ablation study. The title under the image denotes the de-identification setting, and the Original is the original image.}
	\label{fig:ablation}
\end{figure}

Quantitative results for all experiments are presented in Tab. \ref{tab:de-id_ablation}, Tab. \ref{tab:quality_ablation}, and Tab. \ref{tab:usability_ablation}, which report performance in terms of de-identification accuracy, image quality, and usability. Horizontal lines delineate each experimental setting, and the best result within each segment is highlighted in bold. Visual outcomes are provided in Fig. \ref{fig:ablation}, clearly illustrating the qualitative differences across configurations.

\begin{table}[!t]
	\caption{The de-identification performance of ablation study (\%). The symbols ↑ (↓) indicate that the higher (lower) value is better, and the bold means the best result in the corresponding segment.\label{tab:de-id_ablation}}
	\centering
	\begin{tabular}{c|ccc|ccc}
		\toprule
		\multirow{2}{*}{\textbf{Setting}} & \multicolumn{3}{c|}{\textbf{MTCC}} & \multicolumn{3}{c}{\textbf{CCNet}} \\
		& \textbf{RR↑}    & \textbf{DIR↑}   & \textbf{Acc.↓}  & \textbf{RR↑}     & \textbf{DIR↑}   & \textbf{Acc.↓}  \\
		\midrule
		s                        & 97.54  & 97.00  & 6.65   & 97.23           & 83.64          & 8.77   \\
		m                        & 97.77  & 97.08  & 6.58   & 97.65           & 84.15          & 8.35   \\
		f                        & \textbf{98.00}  & \textbf{97.68}  & \textbf{6.38}   & \textbf{98.23}  & \textbf{85.04} & \textbf{7.81}   \\
		\midrule
		s+m                      & 97.50  & 96.85  & 7.15   & 97.15           & 83.61          & 8.77    \\
		s+f                      & 97.73  & 97.05  & 6.62   & 97.08           & 83.92          & 9.04   \\
		m+f                      & \textbf{97.88}  & \textbf{97.26}  & \textbf{6.54}   & \textbf{98.00}  & \textbf{84.39} & \textbf{8.31}   \\
		s+m+f                    & 97.58  & 96.99  & 6.81   & 97.38           & 83.89          & 8.81   \\
		\midrule
		$\alpha$=0.1                    & \textbf{95.42}  & \textbf{94.79}  & \textbf{10.35}  & \textbf{96.42}  & \textbf{82.55} & \textbf{10.12}   \\
		$\alpha$=0.2                    & 91.46  & 92.50  & 15.46  & 95.65           & 81.45          & 11.81   \\
		$\alpha$=0.3                    & 85.00  & 89.87  & 22.96  & 94.15           & 80.07          & 15.00   \\
		\midrule
		Masking                  & \textbf{100.00} & \textbf{125.31} & \textbf{0.31}   & \textbf{100.00} & \textbf{82.79} & \textbf{0.38}   \\
		Blurring                 & 0.04   & 34.94  & 100.00 & 12.23           & 48.79          & 88.73   \\
		Pixelating               & 3.38   & 44.83  & 97.92  & 13.27           & 50.13          & 94.04   \\
		\bottomrule
	\end{tabular}
\end{table}

\begin{table*}[!t]
	\caption{Image quality of ablation study. The symbols ↑ (↓) indicate that the higher (lower) value is better, and the bold means the best result in the corresponding segment.\label{tab:quality_ablation}}
	\resizebox{1\linewidth}{!}{
	\centering
	\begin{tabular}{c|ccccc}
		\toprule
		\textbf{Dataset} & \textbf{SSIM↑}         & \textbf{MS-SSIM↑}      & \textbf{PSNR↑}          & \textbf{LPIPS↓}        & \textbf{FID↓}   \\
		\midrule
		s          & \textbf{0.9632±0.0105} & \textbf{0.9620±0.011}  & \textbf{32.4559±2.3064} & \textbf{0.1069±0.0253} & 11.3220  \\
		m          & 0.9622±0.0107 & 0.9604±0.0113 & 32.038±2.443   & 0.1070±0.0254 & \textbf{6.7847}   \\
		f          & 0.9615±0.011  & 0.9591±0.0119 & 31.6397±2.904  & 0.1076±0.0255 & 8.0160   \\
		\midrule
		s+m        & \textbf{0.9629±0.0106} & \textbf{0.9614±0.0111} & \textbf{32.2961±2.349}  & \textbf{0.1068±0.0253} & 7.4340   \\
		s+f        & 0.9628±0.0106 & 0.9612±0.0111 & 32.2396±2.3626 & 0.107±0.0253  & 7.4082   \\
		m+f        & 0.9621±0.0107 & 0.96±0.0114   & 31.9792±2.4721 & 0.1072±0.0254 & \textbf{6.7122}   \\
		s+m+f      & 0.9626±0.0106 & 0.961±0.0112  & 32.1944±2.3704 & 0.1069±0.0253 & 6.9785   \\
		\midrule
		$\alpha$=0.1      & 0.9633±0.0104 & 0.9628±0.0106 & 32.656±2.2638  & 0.1064±0.0252 & 6.8021   \\
		$\alpha$=0.2      & 0.9636±0.0102 & 0.9638±0.0102 & 32.8863±2.1835 & 0.1061±0.0252 & \textbf{6.7701}   \\
		$\alpha$=0.3      & \textbf{0.9637±0.0102} & \textbf{0.9647±0.0097} & \textbf{33.0864±2.1076} & \textbf{0.1058±0.0251} & 6.8831   \\
		\midrule
		Masking    & 0.9145±0.0184 & 0.9381±0.0115 & 13.9204±1.141  & 0.0764±0.0159 & 168.0455 \\
		Blurring   & \textbf{0.9873±0.0038} & \textbf{0.9852±0.0047} & \textbf{41.0394±1.7601} & \textbf{0.0562±0.012}  & \textbf{30.7098}  \\
		Pixelating & 0.9794±0.0054 & 0.9753±0.0067 & 37.5774±1.6161 & 0.0626±0.0134 & 106.6842 \\
		\bottomrule
	\end{tabular}
}
\end{table*}

\begin{table*}[!t]
	\caption{Usability differences of ablation study ↓ (\%). The range of the value is [0-100]\%, and the lower value is the better. The bold means the best result in the corresponding segment.\label{tab:usability_ablation}}
	\resizebox{1\linewidth}{!}{
	\centering
	\begin{tabular}{c|ccccc}
		\toprule
		\textbf{Dataset} & \makecell{\textbf{Hand} \\ \textbf{Detection}} & \makecell{\textbf{2D Keypoint} \\ \textbf{Detection}} & \makecell{\textbf{3D Keypoint} \\ \textbf{Detection}} & \makecell{\textbf{ROI} \\ \textbf{Localization}} & \makecell{\textbf{Hand} \\ \textbf{Segmentation}} \\
		\midrule
		s          & 0.26±0.26   & 0.274±0.117 & 0.203±0.12  & 0.23±0.872   & 0.251±0.384  \\
		m          & \textbf{0.247±0.235} & \textbf{0.262±0.101} & \textbf{0.185±0.107} & \textbf{0.214±0.8}    & \textbf{0.229±0.373}  \\
		f          & 0.25±0.237  & 0.263±0.112 & 0.186±0.109 & 0.513±2.067  & 0.383±1.022  \\
		\midrule
		s+m        & 0.245±0.246 & 0.262±0.1   & 0.189±0.109 & 0.171±0.431  & 0.215±0.303  \\
		s+f        & 0.247±0.251 & 0.26±0.1    & 0.187±0.107 & 0.173±0.439  & 0.216±0.298  \\
		m+f        & 0.246±0.235 & 0.26±0.103  & 0.186±0.171 & 0.226±0.894  & 0.235±0.42   \\
		s+m+f      & \textbf{0.244±0.244} & \textbf{0.259±0.099} & \textbf{0.185±0.107} & \textbf{0.17±0.438}   & \textbf{0.209±0.257}  \\
		\midrule
		$\alpha$=0.1      & 0.235±0.223 & 0.258±0.096 & 0.186±0.108 & 0.146±0.255  & 0.182±0.151  \\
		$\alpha$=0.2      & \textbf{0.233±0.22}  & 0.257±0.096 & 0.185±0.106 & 0.142±0.307  & 0.172±0.105  \\
		$\alpha$=0.3      & 0.238±0.226 & \textbf{0.253±0.095} & \textbf{0.184±0.106} & \textbf{0.138±0.132}  & \textbf{0.168±0.084}  \\
		\midrule
		Masking    & 2.335±1.193 & 0.488±0.173 & 0.334±0.364 & 10.324±3.607 & 10.488±2.018 \\
		Blurring   & \textbf{0.447±0.329} & 0.355±0.118 & 0.397±0.157 & \textbf{0.116±0.13 }  & \textbf{0.14±0.093}   \\
		Pixelating & 0.558±0.467 & \textbf{0.231±0.093} & \textbf{0.156±0.091} & 0.251±0.524  & 0.39±0.376   \\
		\bottomrule
	\end{tabular}
}
\end{table*}

In Part A of our experiments, using the full ROI (f) led to the highest de-identification performance and lowest image quality, likely due to the semantic noise and distortion brought from its coverage of palmprint-unrelated regions. However, the small ROI (s) delivered the best image quality by preserving finer textures and minimizing visual distortion. The medium ROI (m) emerged as the most balanced in terms of usability, introducing minimal perceptual disruption while still achieving reasonable anonymization. These findings underscore the trade-offs inherent in choosing the ROI scale, with each offering distinct advantages.

Part B explored the fusion of multiple ROI scales to enhance semantic richness. Among the tested combinations, the fusion of medium and full ROIs (m+f) yielded the strongest de-identification performance and worst image quality, likely due to the semantic noise and distortion generally existing in both full ROI (f) and medium ROI (m). In contrast, the small and medium ROI combination (s+m) achieved the best image quality, producing outputs that were visually coherent and natural. Notably, the fusion of all three ROIs (s+m+f) resulted in the smallest usability difference, suggesting that multi-scale integration facilitates effective anonymization with minimal impact on downstream tasks. Balancing these outcomes, the s+m strategy was selected as the optimal SGE fusion approach, as it consistently delivered strong performance across all evaluation metrics while preserving visual realism.

Part C focused on tuning the \textit{prior interpolation} factor $\alpha$ in the latent conditioning map under the s+m fusion strategy. When $\alpha$ was set to 0.1, the framework achieved its strongest de-identification capability, effectively suppressing identifiable traits. As $\alpha$ increases, the interpolated latent moves closer to the original image, incorporating more prior knowledge into the generation process. As a result, the image quality and usability become more similar to the original image, but anonymization strength is correspondingly reduced. Based on these observations, $\alpha$ = 0.1 was identified as the most balanced setting, offering robust de-identification while avoiding visually implausible artifacts. This value was adopted as the default interpolation factor in our experiments.

To further validate the robustness of our method, Part D compared it against traditional anonymization techniques, including masking, blurring, and pixelation. While masking completely removed the palmprint region, thereby delivering perfect de-identification, it came at the cost of severely degraded image quality and usability, rendering it unsuitable for practical use. On the other hand, blurring and pixelation preserved more of the original image structure but failed to anonymize the biometric content effectively. Their poor de-identification performance rendered any perceived gains in visual quality or usability irrelevant.

Collectively, these experiments validate the effectiveness and adaptability of our framework, demonstrating its ability to produce high-quality, privacy-preserving image generation without compromising usability, a distinct advantage over conventional anonymization methods.

\section{Discussion}\label{sec:5}
We adopt a training-free, optimization-free design for three key reasons: privacy and governance (avoiding fine-tuning on biometric corpora minimizes the risk of model memorization and avoids consent and data-retention requirements), cross-domain generality (an off-the-shelf generator reduces overfitting to specific datasets or capture protocols), and reproducibility cost (results can be reproduced without large palmprint datasets or extensive computation). While this approach may slightly compromise modality-specific texture fidelity, it ensures stronger privacy and broader applicability. Importantly, the proposed plug-and-play mechanisms—SGE fusion and prior interpolation—provide flexibility, allowing the framework to be readily adjusted or reconfigured for different scenarios simply by tuning guidance or parameters, without any additional training or data collection. Overall, this paradigm achieves a practical balance between privacy, robustness, and accessibility, while leaving room for principled, privacy-aware adaptations in future work.

\section{Conclusions and Future Works}\label{sec:6}
This paper presents the first palmprint de-identification framework that jointly addresses de-identification effectiveness, visual quality, and usability, thereby establishing a balanced baseline for this task. Our approach operates in a training-free and optimization-free fashion, making it both efficient and adaptable. By leveraging SGE fusion and \textit{prior interpolation} within a pre-trained diffusion model, the method enables stable and controllable identity obfuscation. Experimental results demonstrate that our method achieves strong de-identification performance, high image fidelity, inherent diversity across outputs, and minimal impact on downstream usability. Nevertheless, there remains significant room for enhancing de-identification robustness, particularly under more challenging conditions. Future research will extend this framework to full-hand and multi-modal biometric de-identification scenarios, further broadening its applicability.

\section{Acknowledgments}
This work was supported by the Science and Technology Development Fund, Macao S.A.R (FDCT) [grant number  0028/2023/RIA1]; the National Natural Science Foundation of China [grant number 62466038] and the Jiangxi Provincial Key Laboratory of Image Processing and Pattern Recognition [grant number 2024SSY03111].

\bibliographystyle{elsarticle-num} 
\bibliography{reference}



%\begin{thebibliography}{00}
%
%%% For numbered reference style
%%% Text of bibliographic item
%
%\bibitem{lamport94}
%  \textit{\LaTeX: a document preparation system},
%  Addison Wesley, Massachusetts,
%\end{thebibliography}
\end{document}